\documentclass[10pt,twocolumn,letterpaper]{article}

\usepackage{cvpr}
\usepackage{times}
\usepackage{epsfig}
\usepackage{graphicx}
\usepackage{amsmath}
\usepackage{amssymb}

\usepackage[pagebackref=true,breaklinks=true,letterpaper=true,colorlinks,bookmarks=false]{hyperref}

\usepackage{tabularx}
\usepackage{booktabs}
\usepackage{epstopdf}
\usepackage{algorithm}
\usepackage{algorithmic}
\usepackage{threeparttable}
\newtheorem{theorem}{Theorem}
\usepackage{array}
\newcommand{\PreserveBackslash}[1]{\let\temp=\\#1\let\\=\temp}
\newcolumntype{C}[1]{>{\PreserveBackslash\centering}p{#1}}
\newcolumntype{R}[1]{>{\PreserveBackslash\raggedleft}p{#1}}
\newcolumntype{L}[1]{>{\PreserveBackslash\raggedright}p{#1}}

\usepackage[english]{babel}

\cvprfinalcopy 

\ifcvprfinal\fi
\begin{document}

\title{Deep Discriminative Clustering Analysis}

\author{Jianlong Chang$^{1,2}$~~Yiwen Guo$^{3}$~~Lingfeng Wang$^{1}$~~Gaofeng Meng$^{1}$~~Shiming Xiang$^{1,2}$~~Chunhong Pan$^{1}$\\
$^1$ National Laboratory of Pattern Recognition, Institute of Automation, Chinese Academy of Sciences\\
$^2$ School of Artificial Intelligence, University of Chinese Academy of Sciences\\
$^3$ Intel Labs China\\
{\tt\small \{jianlong.chang, lfwang, gfmeng, smxiang, chpan\}@nlpr.ia.ac.cn}~~~\tt\small yiwen.guo@intel.com\\
}

\maketitle

\begin{abstract}
Traditional clustering methods often perform clustering with low-level indiscriminative representations and ignore relationships between patterns, resulting in slight achievements in the era of deep learning. To handle this problem, we develop Deep Discriminative Clustering (DDC) that models the clustering task by investigating relationships between patterns with a deep neural network. Technically, a global constraint is introduced to adaptively estimate the relationships, and a local constraint is developed to endow the network with the capability of learning high-level discriminative representations. By iteratively training the network and estimating the relationships in a mini-batch manner, DDC theoretically converges and the trained network enables to generate a group of discriminative representations that can be treated as clustering centers for straightway clustering. Extensive experiments strongly demonstrate that DDC outperforms current methods on eight image, text and audio datasets concurrently.
\end{abstract}


\section{Introduction}

Clustering is one of the most extensively used statistical tool in computer vision and pattern recognition \cite{Achanta_2017_CVPR,Agudo_2018_CVPR,chang2018deep,Chang2017PR,DBLP:conf/cvpr/JoulinBP10,Liu_2018_CVPR}. In a wide range of fields, many applications where the target information is unobservable can be intrinsically instantiated as clustering tasks, including text mining \cite{Lin_2018_CVPR} and multimedia content-based retrieval \cite{DBLP:conf/cvpr/AchantaS17,DBLP:conf/iccv/DizajiHDCH17,Zhang_2017_CVPR,Zhao_2018_CVPR}.

In the literature, much research has been dedicated to clustering analysis \cite{DBLP:conf/icml/BhaskaraW18, DBLP:conf/icml/ChatziafratisNC18,DBLP:conf/iccv/0004HNCW17,Ikami_2018_CVPR, Lin_2018_CVPR}. From the aspect related to algorithmic structure and operation, clustering methods can be roughly divided into two categories, \textit{i.e.}, partitional and agglomerative methods \cite{DBLP:journals/csur/JainMF99}. By definition, partitional methods tend to split all patterns into several partitions, until stopping criterions are met. In compliance with this definition, typically basic methods are present, such as the Kmeans \cite{DBLP:journals/tkde/WangWSXSL15} and the spectral clustering \cite{DBLP:conf/nips/Zelnik-ManorP04}. In contrast, the agglomerative methods \cite{DBLP:conf/icml/ChatziafratisNC18,DBLP:journals/pami/FrantiVH06} always start with each pattern as a single cluster and aggregate clusters together gradually until stopping criterions are satisfied. Despite the conceptual breakthroughs, these clustering methods suffer from a severe obstacle, \textit{i.e.}, how to acquire effective representations from unlabeled data purely?

In order to acquire effective representations, unsupervised techniques have been developed to represent data. Handcrafted representations, derived from the professional knowledge, are originally utilized, such as SIFT \cite{DBLP:conf/iccv/Lowe99} and HoG \cite{DBLP:conf/cvpr/DalalT05}. However, they are always limited to simple scenarios, and may be severely degenerated when facing with more complex ones, including large scale images, texts, and audios. By employing deep unsupervised learning techniques \cite{DBLP:conf/nips/BengioLPL06,DBLP:journals/jmlr/VincentLLBM10}, more preferable representations can be acquired and considerable gains are achieved consequently. Unfortunately, these learned representations are fixed in clustering and can not be further modified to yield better performance. Analogous to previous studies in deep supervised leaning, learnable representations are usually more effective than fixed ones. To benefit from learnable representations, one straightforward way is to integrate deep representation learning and clustering in a joint framework \cite{Chang2017ICCV,DBLP:journals/corr/abs-1803-01449,DBLP:conf/icml/XieGF16,DBLP:conf/cvpr/YangPB16}, but there are still several challenges. First, how to define an effective and general objective to train deep networks in an unsupervised way? Second, how to guarantee that the representations learned by deep networks are in favour of clustering? Third, how to discover the number of clusters automatically, rather than to predefine one.

To address such challenges, we propose Deep Discriminative Clustering (DDC) that manages the clustering task by iteratively exploring relationships between patterns and learning representations in a mini-batch manner. In each iteration, a global constraint is used to guide the estimation of the relationships. Then under a local constraint, the relationships are fed back to train the network for learning high-level discriminative representations. Consequently, DDC is theoretically convergent and the trained network is capable of generating a group of discriminative representations that can be treated as clustering centers for straightway clustering. Benefiting from such artful modeling, DDC is independent of the number of patterns and the number of clusters, which endows the model with the capability of dealing with tasks that with a large number of patterns or unobservable number of clusters.


To sum up, the key contributions of this work are:
\begin{itemize}
  \item Under the global and local constraints, DDC can endow networks with the capability of mapping all patterns to discriminative clustering centers for clustering.
  \item By feat of the mini-batch based optimization, DDC alleviates the fussy impacts from the numbers of patterns and clusters, further enhancing the practicality.
  \item The availability and the convergence of the developed DDC model are mathematically analyzed, which provides requisite theoretical foundations and guarantees.
  \item Extensive experiments strongly verify that our DDC model is concurrently superior to current methods on various datasets, including images, texts, and audios.
\end{itemize}

\section{Related Work}

In this section, we make a brief review of the related work on the clustering and deep learning methods.

\subsection{Clustering}

By definition, clustering tries to generate a semantical organization of data based on similarities, namely patterns within the same cluster are similar to each other, while those in different clusters are dissimilar. Technically, clustering has been widely studied in an unsupervised mechanism from two main aspects: how to define a cluster and how to extract effective representations for clustering?

To define clusters, lots of endeavors have been extensively made. In particular, the Gaussian mixture distribution is one of the simple yet popular techniques to describe clusters. Relying on this technique, a series of clustering methods are developed, such as the Kmeans \cite{DBLP:journals/tkde/WangWSXSL15} and its variants \cite{DBLP:conf/icml/Sinha18,DBLP:journals/tkde/WangWSXSL15,DBLP:conf/nips/YeZW07}. Compared with those with predefined clusters, many methods prefer to adaptively estimate forms of clusters from datasets purely. Typically, there are two frequently-used ways, \textit{i.e.}, density-based and connectivity-based estimations. For the former, density-based methods define clusters via a proper density function, which can be utilized to describe the density around each patterns in the feature space. Depending on diverse density functions, various density-based methods\cite{marin2017kernel, rodriguez2014clustering,DBLP:conf/icmlc/WangWTH15} have been proposed. For the latter, patterns are highly connected if they belong to the same cluster, including the spectral clustering \cite{DBLP:conf/icml/BajajGHHL18,Wang_2017_CVPR}. Furthermore, these ideas also form the basis of a number of methods \cite{DBLP:conf/icml/ChenQ16,DBLP:journals/jmlr/YangCO16}, such as the discriminative clustering \cite{DBLP:conf/nips/GomesKP10,DBLP:conf/iccv/LuoHD11, DBLP:conf/iccv/MiechABLS17,DBLP:conf/nips/MoosmannTJ06, DBLP:conf/icml/TorreK06,DBLP:conf/iccv/Tu05a, DBLP:conf/cvpr/XuHN16, DBLP:conf/nips/YeZW07,DBLP:conf/cvpr/ZografosEM13}, and the subspace based clustering \cite{DBLP:conf/cvpr/LiV15,Peng_2017_CVPR,Sznaier_2018_CVPR,DBLP:conf/cvpr/YinGGHX16,DBLP:conf/cvpr/YouRV16,Zhou_2018_CVPR}.

To extract effective representations for clustering, plentiful attention has been paid. Primitively, many handcrafted representations have been employed to recode patterns in an unsupervised way, including SIFT \cite{DBLP:conf/iccv/Lowe99} and HoG \cite{DBLP:conf/cvpr/DalalT05} on images. To eliminate influences from trivial variables (\textit{e.g.}, variations of scenes and objects), high-level representations can be learned in a data-driven manner with deep unsupervised learning in recent \cite{DBLP:conf/nips/BengioLPL06,DBLP:journals/corr/DilokthanakulMG16,DBLP:journals/corr/Springenberg15, DBLP:journals/jmlr/VincentLLBM10,DBLP:conf/cvpr/ZeilerKTF10, DBLP:journals/corr/ZhaoMGL15}.

In spite of the remarkable successes \cite{Gholami_2017_CVPR, DBLP:conf/icml/KamnitsasCFWTRG18,DBLP:conf/icml/LangeKA18,DBLP:conf/icml/MartinLV18, Zhang_2017_CVPR}, these two aspects are always separated. As a result, the extracted representations are fixed during clustering and can not be further improved to obtain better performance.

\subsection{Deep Neural Network}

Deep Neural Network (DNN) is a significant technique in machine learning, which has been developed to manage tasks like humans do \cite{DBLP:journals/nature/LeCunBH15}. With respect to the observability of label information, deep learning can be summarized into two categories, \textit{i.e.}, deep supervised learning and deep unsupervised learning.

In deep supervised learning, much interesting research has been proposed over the last decade. Thanks to the efficient hardware and a large amount of labeled data, remarkable successes have been achieved in various pattern analysis and machine learning problems, especially in computer vision, such as image classification and objection detection \cite{Chang_2018_NIPS,girshick2014rich,DBLP:journals/corr/HeZRS15,DBLP:conf/iccv/HeZRS15, long2015fully,szegedy2015going}. Although the achievements are significant, deep supervised learning solely pertains to the tasks that with a large amount of labeled data.

To eliminate the requirement for the labeled data, deep unsupervised learning has drawn attention to train DNNs in an unsupervised manner and extract representations of data simultaneously \cite{Caron_2018_ECCV,Mi_2018_ECCV,DBLP:journals/corr/RadfordMC15,DBLP:journals/corr/abs-1712-07788}. In general, the vital problem in deep unsupervised learning is that how to definitely give pseudo labels to train DNNs. Primitively, most unsupervised learning methods attempt to reconstruct inputs with DNNs, namely labels are themselves. Typical approaches include the autoencoder \cite{DBLP:conf/nips/BengioLPL06} and its variants \cite{DBLP:journals/corr/MakhzaniSJG15, DBLP:journals/jmlr/VincentLLBM10}. Similar to traditional machine learning methods, more robust and interpretable representations can be learned by introducing additional regularization terms. In recent, deep generative networks \cite{DBLP:conf/nips/GoodfellowPMXWOCB14}, one important branch in deep unsupervised learning has been developed. They train networks by representing data distributions with DNNs. Akin to traditional methods, representations can be learned by modifying generative networks, such as the generative adversarial clustering \cite{DBLP:conf/ijcai/JiangZTTZ17,DBLP:journals/corr/abs-1804-11130,DBLP:journals/corr/abs-1809-03627,premachandran2016unsupervised, DBLP:journals/corr/Springenberg15} and the variational autoencoder based clustering \cite{DBLP:journals/corr/DilokthanakulMG16,DBLP:journals/corr/MakhzaniSJG15}.

While many endeavors have been devoted to deep unsupervised learning, several intrinsic challenges still remain. First, how to provide proper driving force to train DNNs in an unsupervised manner? Second, clustering results can not be obtained directly via the learned feature representations by these methods, which implies that the generated representations are unsuitable for clustering naturally.

\begin{figure*}[!t]
\centering
\centerline{\includegraphics[width=17.5cm]{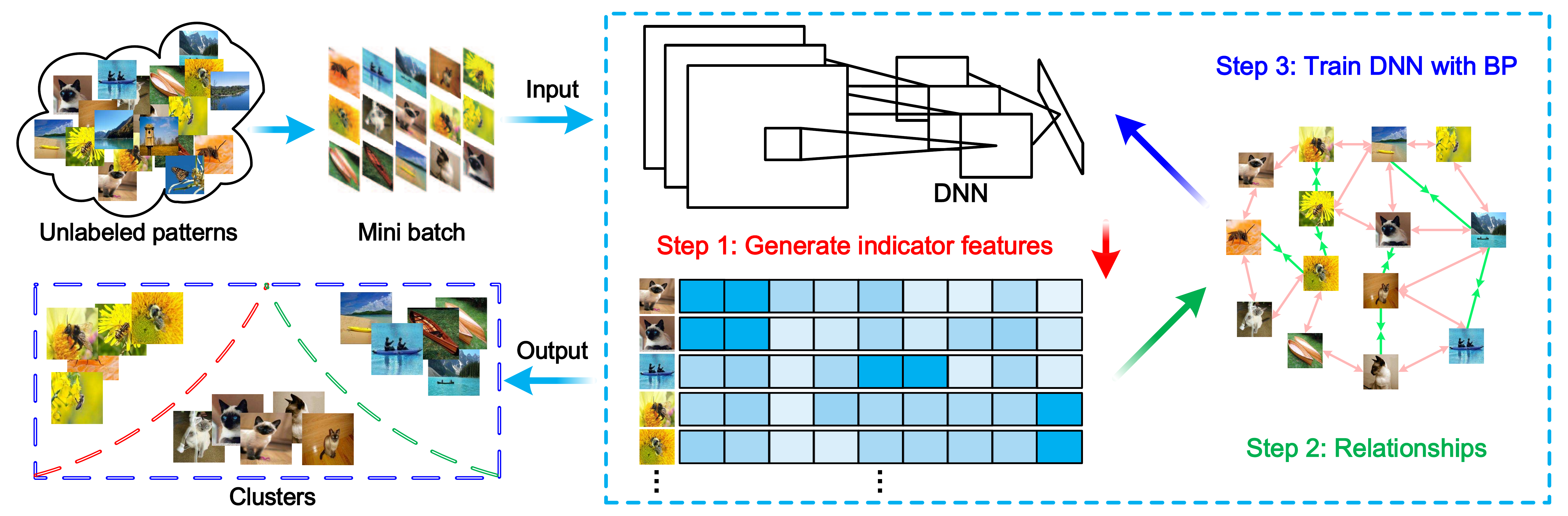}}
\caption{The flowchart of DDC. In each iteration, DDC consists of three essential steps. First, a DNN is built to generate indicator features on a mini batch of patterns. Second, the relationships between patterns are estimated based on the generated indicator features. Third, the DNN is trained by optimizing the relationships, \textit{i.e.}, increasing the similarities of similar patterns (shown in green arrows) and reducing the similarities of dissimilar patterns (shown in red arrows). Iterating step 1 to step 3 until the DDC model can not be further improved. Finally, the trained DNN enables to generate discriminative indicator features that can be treated as clustering centers for clustering directly.
}\medskip
\label{main}
\vspace{-0.45cm}
\end{figure*}

\section{Deep Discriminative Clustering Model}

Clustering, intrinsically, is a function that is in a position to assign patterns into a group of clusters whose number is either observable or unobservable. Formally, considering the problem of clustering $n$ patterns $\mathcal{D}=\{\mathbf{x}_{i}\}_{i=1}^{n}$ into $k$ clusters, \textit{i.e.}, $\mathcal{D}=\mathcal{D}_{1}\bigcup\cdots\bigcup\mathcal{D}_{k}$, which are represented by $k$ discriminative clustering centers $\mathcal{C}=\{\mathbf{c}_{1},\cdots,\mathbf{c}_{k}\}$. By this definition, clustering is naturally formulated as
\begin{equation}
\begin{aligned}\label{clustering_form}
&h(\mathbf{x}_{i};\mathbf{w})\in\mathcal{C},
\end{aligned}
\end{equation}
where $h(\cdot;\mathbf{w})$ is a clustering function with the learnable parameter $\mathbf{w}$. In clustering, generally, a fundamental problem is how to model an objective function on a set of clustering centers $\mathcal{C}$ to estimate the parameter $\mathbf{w}$. In the following, we focus on dealing with this problem. The flowchart of DDC is visually illustrated in Figure~\ref{main}.

\subsection{Relational Objective Function}


The main obstacle of modeling an objective function in clustering is that the correspondence between patterns and clustering centers $\mathcal{C}=\{\mathbf{c}_{1},\cdots,\mathbf{c}_{k}\}$ is non-unique. To eliminate this obstacle, relationships between patterns are employed as the label information, since whether pairwise patterns belong to the same cluster is unique. From this perspective, we reform the dataset $\mathcal{D}=\{\mathbf{x}_{i}\}_{i=1}^{n}$ as a relational dataset $\mathcal{R}=\{(\mathbf{x}_{i},\mathbf{x}_{j},R_{ij})\}_{i=1,j=1}^{n}$, where $R_{ij}=1$ represents that $\mathbf{x}_{i}$ and $\mathbf{x}_{j}$ belong to the same cluster, and $R_{ij}=0$ otherwise. Accordingly, the objective function in clustering can be reformulated as
\begin{equation}
\begin{aligned}\label{rijg(xi,xj)initial}
\min\limits_{\mathbf{w}}\mathbf{E}=\sum_{i,j}\ell(R_{ij},g(\mathbf{x}_{i},\mathbf{x}_{j};\mathbf{w})),
\end{aligned}
\end{equation}
where $g(\cdot,\cdot;\mathbf{w})$ is a learnable function to estimate similarities between patterns, and $\ell(\cdot,\cdot)$ is the binary cross-entropy to measure the error between $R_{ij}$ and $g(\mathbf{x}_{i},\mathbf{x}_{j};\mathbf{w})$. In Eq.~(\ref{rijg(xi,xj)initial}), two issues need to be handled. First, the ground-truth relationship $R_{ij}$ between $\mathbf{x}_{i}$ and $\mathbf{x}_{j}$ is missing in clustering. Second, clustering labels of $\mathbf{x}_{i}$ and $\mathbf{x}_{j}$ can not be explicitly acquired even though the ground-truth relationship $R_{ij}$ is observable. In the following, we manage these issues by introducing a global constraint and a local constraint.

\subsubsection{Global Constraint}

To tackle these issues, we first explore an inherent priori knowledge termed as global constraint in clustering, which can be employed to instantiate and refine the basic model in Eq.~(\ref{rijg(xi,xj)initial}). Specifically, the global constraint naturally models the relationships between patterns with three essential rules, \textit{i.e.}, reflexivity, symmetry and transitivity.

\textbf{Reflexivity} The reflexivity signifies that every pattern $\mathbf{x}_{i}$ and itself strictly belong to the same cluster, \textit{i.e.}, $R_{ii}=1$ is met for arbitrary $i$. Under this rule, we have
\begin{equation}
\begin{aligned}
g(\mathbf{x}_{i},\mathbf{x}_{i};\mathbf{w})=1
\end{aligned}.
\end{equation}
Intrinsically, the reflexivity can be considered as a ``reconstruction'' constraint for every pattern. Contrary to the reconstruction in AE \cite{DBLP:conf/nips/BengioLPL06} that attempts to learn features to represent inputs, the reflexivity focuses on modeling the similar relationship in a high-level latent space.

\textbf{Symmetry} The symmetry indicates that $\mathbf{x}_{i}$ is (dis) similar to $\mathbf{x}_{j}$ if and only if $\mathbf{x}_{j}$ is (dis) similar to $\mathbf{x}_{i}$, \textit{i.e.}, $R_{ij}=R_{ji}$ is met for arbitrary $i$ and $j$. In order to meet this rule, the function $g(\mathbf{x}_{i},\mathbf{x}_{j};\mathbf{w})$ should be a symmetric function for the inputs $\mathbf{x}_{i}$ and $\mathbf{x}_{j}$. For this purpose, the function $g(\cdot,\cdot;\mathbf{w})$ is decomposed into the dot product between the same function $ f(\cdot;\mathbf{w})$, \textit{i.e.},
\begin{equation}
\begin{aligned}\label{symmetry function}
g(\mathbf{x}_{i},\mathbf{x}_{j};\mathbf{w})= f(\mathbf{x}_{i};\mathbf{w})\cdot f(\mathbf{x}_{j};\mathbf{w})
\end{aligned},
\end{equation}
where ``$\cdot$'' represents the dot product. Generally, a deep network is employed to parametrize $ f(\cdot;\mathbf{w})$ because of its recognized capability in feature learning.

\textbf{Transitivity} The transitivity means that $\mathbf{x}_{i}$ is similar to $\mathbf{x}_{h}$ if $\mathbf{x}_{i}$ is similar to $\mathbf{x}_{j}$ and $\mathbf{x}_{j}$ is similar to $\mathbf{x}_{h}$, \textit{i.e.}, $R_{ih}=1$ if $R_{ij}=1=R_{jh}$ for arbitrary $i$, $j$ and $h$. Naturally, the transitivity signifies that the ground-truth similarity matrices in clustering possess a specific form, although the real similarities are unobservable.

Inspired by the above three rules, we attempt to estimate the ground-truth relationships from coarse relationships. In detail, given a deep network $f(\cdot;\mathbf{w})$, the coarse similarity matrix is first obtained according to Eq.~(\ref{symmetry function}). Then under the global constraint, the ground-truth similarity matrix is estimated by finding the most similar matrix to the coarse similarity matrix, \textit{i.e.},
\begin{equation}
\begin{aligned}
&\min\limits_{\mathbf{R}}\varepsilon(\mathbf{R},\mathbf{\bar{R}})=\sum_{i,j}|R_{ij}-\bar{R}_{ij}|\\
&\text{~s.t.~}\mathfrak{g}(\mathbf{R})=True\\
&~~~\forall~i,~j,~\bar{R}_{ij}= f(\mathbf{x}_{i};\mathbf{w})\cdot f(\mathbf{x}_{j};\mathbf{w}),
\end{aligned}
\end{equation}
where $\mathbf{\bar{R}}$ is the coarse similarity matrix calculated with the network $ f(\cdot;\mathbf{w})$, ``$\mathfrak{g}(\mathbf{R})=True$'' implies that $\mathbf{R}$ mets the above global constraint, and $\varepsilon(\mathbf{R},\mathbf{\bar{R}})$ is a global constraint error to measure the error between $\mathbf{R}$ and $\mathbf{\bar{R}}$. By introducing the global constraint into Eq.~(\ref{rijg(xi,xj)initial}), the basic DDC model can be rewritten as follows:
\begin{equation*}
\begin{aligned}
&\min\limits_{\mathbf{w},\mathbf{R}}\mathbf{E}=\sum_{i,j}\ell(R_{ij}, f(\mathbf{x}_{i};\mathbf{w})\cdot f(\mathbf{x}_{j};\mathbf{w}))+\varepsilon(\mathbf{R},\mathbf{\bar{R}})\\
\end{aligned}
\end{equation*}
\vspace{-0.2cm}
\begin{equation}
\begin{aligned}\label{rijg(xi,xj)middle}
&\text{s.t.~}\mathfrak{g}(\mathbf{R})=True\\
&~~\forall~i,~j,~\bar{R}_{ij}= f(\mathbf{x}_{i};\mathbf{w})\cdot f(\mathbf{x}_{j};\mathbf{w}),~~~~~~~\\
\end{aligned}
\end{equation}
where the first term is employed to learn the parameter $\mathbf{w}$ for estimating faithful similarities, and the second term $\varepsilon(\mathbf{R},\mathbf{\bar{R}})$ acts as a penalty to explore the ground-truth relationships from coarse relationships.

\subsubsection{Local Constraint}


To learn beneficial representations for clustering, we introduce indicator features $\mathcal{I}=\{\mathbf{I}_{i}= f(\mathbf{x}_{i};\mathbf{w})\}_{i=1}^{n}$, which are capable of assigning each pattern to a cluster automatically. To this end, we impose a non-negative local constraint on every element in indicator features, \textit{i.e.},
\begin{equation}
\begin{aligned}\label{g(xi,xj)}\forall~i, I_{ih}\geq0,h=1,2,\cdots,k
\end{aligned},
\end{equation}
where $I_{ih}$ signifies the $h$-th element in the indicator feature $\mathbf{I}_{i}\in\mathbb{R}^{k}$, and $\mathbf{I}_{i}= f(\mathbf{x}_{i};\mathbf{w})$ is satisfied. Compared with the global constraint that focuses on the relationships between patterns, the local constraint is only concerned with local information, \textit{i.e.}, each element in indicator features. By integrating Eq.~(\ref{g(xi,xj)}) into Eq.~(\ref{rijg(xi,xj)middle}), the objective function of DDC is formulated as follows:
\begin{equation*}
\begin{aligned}
&\min\limits_{\mathbf{w},\mathbf{R}}\mathbf{E}=\sum_{i,j}\ell(R_{ij}, f(\mathbf{x}_{i};\mathbf{w})\cdot f(\mathbf{x}_{j};\mathbf{w}))+\varepsilon(\mathbf{R},\mathbf{\bar{R}})\\
\end{aligned}
\end{equation*}
\vspace{-0.2cm}
\begin{equation}
\begin{aligned}\label{rijg(xi,xj)final}
&\text{s.t.~}\mathfrak{g}(\mathbf{R})=True\\
&~~\forall~i, I_{ih}\geq0,h=1,2,\cdots,k\\
&~~\forall~i,~j,~\bar{R}_{ij}= f(\mathbf{x}_{i};\mathbf{w})\cdot f(\mathbf{x}_{j};\mathbf{w}).~~~~~~~\\
\end{aligned}
\end{equation}

In the following, we give Theorem \ref{theorem1} and Theorem \ref{theorem2} to theoretically analyze the availability of our DDC model.


\begin{theorem}\label{theorem1}
If the optimal minimum of Eq.~(\ref{rijg(xi,xj)final}) is attained, indicator features are either equivalent or orthogonal, \textit{i.e.},\\
\centerline{$\mathbf{I}_{i}\neq\mathbf{I}_{j}\Leftrightarrow R_{ij}=0$ or $\mathbf{I}_{i}=\mathbf{I}_{j}\Leftrightarrow R_{ij}=1$.}
Further, the locations of the largest responses in indicator features are different if $R_{ij}=0$, and $R_{ij}=1$ otherwise.
\end{theorem}
\noindent Theorem \ref{theorem1} indicates that the deep network $ f(\cdot;\mathbf{w})$ in DDC intrinsically learns a group of discriminative clustering centers. Since different locations of the largest response are uniquely corresponding to different clustering centers, all patterns can be grouped by locating the largest response in indicator features, \textit{i.e.},
\begin{equation}
\begin{aligned}\label{labeling}
c_{i}=\arg\max_{h}\left(I_{ih}\right)
\end{aligned},
\end{equation}
where $c_{i}$ and $\mathbf{I}_{i}$ are the cluster label and the learned indicator feature of $\mathbf{x}_{i}$, respectively. We also provide a sufficient condition for achieving the global minimum as follows.

\begin{theorem}\label{theorem2}
For an arbitrary similarity matrix $\mathbf{R}$ with $k$ clusters, the global minimum of Eq.~(\ref{rijg(xi,xj)final}) can be always attained via a powerful enough function $ f(\cdot;\mathbf{w})$, if only\\
\centerline{the dimension of indicator features is not smaller than $k$.}
\end{theorem}
\noindent Theorem \ref{theorem2} implies that DDC enables to handle a set of unlabeled data with $k$ clusters if only the dimension of indicator features is not smaller than $k$. This obviously expands the practicability of DDC, since it is unnecessary to predefine the number of clusters $k$ before clustering.

\subsection{In Context of Related Work}

Our DDC model is established by introducing the local and global constraints into a supervised objective. Generally, the local constraint limits the non-negativity of each element in indicator features, which focuses on the local information merely. In contrast, the global constraint is capable of considering relationships among multiple patterns. In additional, the developed DDC model can be treated as a generalization of the DAC model \cite{Chang2017ICCV}. In the following, we discuss the crucial differences in details.

Technically, DAC enables to learn one-hot representations for clustering. Contrary to our DDC, three practical drawbacks make this work less general and adaptive. First, DAC solely pertains to focus on relationships between pairwise patterns, which is insufficient for clustering. For example, $\mathbf{x}_{i}$ may be dissimilar to $\mathbf{x}_{h}$ in DAC, if $\mathbf{x}_{i}$ is similar to $\mathbf{x}_{j}$ and $\mathbf{x}_{j}$ is similar to $\mathbf{x}_{h}$. Because of the transitivity, DDC is in a position to avoid such a fussy scenario. Second, DAC requires to predefine the number of clusters, which may hardly be observable before clustering. By comparison, DDC can automatically estimate the number of clusters in a purely data-driven way, as demonstrated in Theorem \ref{theorem2}. Third, the performance of DAC is very sensitive to the hyper-parameters for estimating relationships. Relying on the transitivity in the global constraint, in contrast, DDC is competent in estimating relationships without additional hyper-parameters, further enhancing the dependability.

\begin{figure}[!t]
\centering
\centerline{\includegraphics[width=7.5cm]{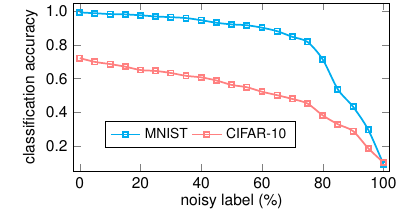}}
\caption{The motivation of the DDC algorithm. The tendencies signify that DNNs are relatively robustness to noisy labels.
}\medskip
\label{motivation}
\vspace{-0.45cm}
\end{figure}

\section{Deep Discriminative Clustering Algorithm}

For optimizing the DDC model, we develop a straightforward yet effective DDC algorithm. Naturally, the algorithm focuses on two problems in Eq.~(\ref{rijg(xi,xj)final}), namely the constraint on indicator features and the optimization.

\subsection{Implementation of Indicator Features}

A constraint layer is built to learn the indicator features to met the non-negativity and the reflexivity in the local and global constraints, respectively. Formally, the layer consists of two mapping functions, \textit{i.e.},
\begin{subequations}
\begin{align}
z_{h}^{t}:&=\exp{\left(z_{h}^{i}-\max\limits_{h}\left(z_{h}^{i}\right)\right)},~h=1,2,\cdots,k,\label{constraint layer_1}\\
z_{h}^{o}:&=\frac{z_{h}^{t}}{\parallel\mathbf{z}^{t}\parallel_{2}},~h=1,2,\cdots,k,\label{constraint layer_2}
\end{align}
\end{subequations}
where $\mathbf{z}^{i}$, $\mathbf{z}^{t}$ and $\mathbf{z}^{o}\in\mathbb{R}^{k}$ indicate the input, temporary variable and output of the layer, respectively. Through the transformation of the above functions, outputs of arbitrary DNNs can be mapped as the indicator features consequently.

\subsection{Alternating Optimization on Mini Batches}

We solve the DDC model by iteratively optimizing the two terms in Eq.~(\ref{rijg(xi,xj)final}) on mini batches of patterns, which is inspired by a discovery in training DNNs. As illustrated in Figure~\ref{motivation}, DNNs have considerable robustness to noisy labels when the rate of noisy label is small \cite{DBLP:conf/cvpr/XieWWWT16}. Intrinsically, the core problem in training DNNs is to find proper gradients on mini batches. If only a few noisy labels are imported, the real gradients may not be severely degenerated. Benefiting from this discovery, the DDC algorithm iteratively updates the $ f(\cdot;\mathbf{w})$ to alleviate the impacts of noises in the estimated similarity matrices. In summary, the algorithm is illustrated in Algorithm~\ref{DDC algorithm} and elaborated as follows.

\begin{algorithm}[t]
\renewcommand{\algorithmicrequire}{\textbf{Input:}}
\renewcommand{\algorithmicensure}{\textbf{Output:}}
\caption{Deep Discriminative Clustering}
\label{DDC algorithm}
\begin{algorithmic}[1]
    \REQUIRE Dataset $\mathcal{D}=\left\{\mathbf{x}_{i}\right\}_{i=1}^{n},~ f(\cdot;\mathbf{w}),~m,~\epsilon$.
    \ENSURE Clustering label $c_{i}$ of $\mathbf{x}_{i}\in \mathcal{D}$.
    \STATE Randomly initialize $\mathbf{w}$;
    \REPEAT
       \FORALL{$q\in\left\{1,2,\cdots,\left\lfloor\frac{n}{m}\right\rfloor\right\}$}\label{begin}
          \STATE Sample batch $\mathcal{D}_{q}$ from $\mathcal{D}$; $//$ $m$ patterns in $\mathcal{D}_{q}$
          \STATE Generate $\mathbf{R}$ on the batch $\mathcal{D}_{q}$; $//$ Eq.~(\ref{rijg(xi,xj)R})
          \STATE Update $\mathbf{w}$ by minimizing Eq.~(\ref{rijg(xi,xj)I});
       \ENDFOR\label{end}
    \UNTIL{$\varepsilon(\mathbf{R},\mathbf{\bar{R}})<\epsilon$}
    \FORALL{$\mathbf{x}_{i}\in \mathcal{D}$}
        \STATE $\mathbf{I}_{i}:= f(\mathbf{x}_{i};\mathbf{w})$;
        \STATE $c_{i}:=\arg\max\nolimits_{h}(I_{ih})$;
    \ENDFOR
\end{algorithmic}
\end{algorithm}

When $\mathbf{w}$ in the function $ f(\cdot;\mathbf{w})$ is fixed, the DDC model in Eq.~(\ref{rijg(xi,xj)final}) degenerates as follows:
\begin{equation}
\begin{aligned}\label{rijg(xi,xj)R}
&\min\limits_{\mathbf{R}}\mathbf{E}=\varepsilon(\mathbf{R},\mathbf{\bar{R}})\\
&\text{~s.t.~}\mathfrak{g}(\mathbf{R})=True;~\bar{R}_{ij}= f(\mathbf{x}_{i};\mathbf{w})\cdot f(\mathbf{x}_{j};\mathbf{w}),
\end{aligned}
\end{equation}
which can be recast as the normalized cut problem on $\mathcal{I}=\{\mathbf{I}_{i}\}_{i=1}$. Therefore, we acquire $\mathbf{R}$ relying on the spectral clustering \cite{DBLP:conf/nips/Zelnik-ManorP04} method. Because of the mini-batch based optimization, the efficiency of our DDC model can be guaranteed for large datasets.

Once the similarity matrix $\mathbf{R}$ is calculated, the DDC model in Eq.~(\ref{rijg(xi,xj)final}) is equivalent to
\begin{equation}
\begin{aligned}\label{rijg(xi,xj)I}
&\min\limits_{\mathbf{w}}\mathbf{E}=\sum_{i,j}\ell(R_{ij},f(\mathbf{x}_{i};\mathbf{w})\cdot f(\mathbf{x}_{j};\mathbf{w}))\\
&\text{s.t.~}\mathfrak{g}(\mathbf{R})=True;~I_{ih}\geq0,h=1,2,\cdots,k,\\
\end{aligned}
\end{equation}
which is a typical supervised task, and can be solved with the back-propagation algorithm.

Synthetically, the convergence of the DDC algorithm is guaranteed in the following Theorem~\ref{theorem3}, \textit{i.e.},
\begin{theorem}\label{theorem3}
If the sampled mini batches in DDC are identically distributed, the DDC algorithm is convergent.
\end{theorem}
\noindent Since the assumption in Theorem~\ref{theorem3} is always satisfied in general scenarios, the DDC model is convergent.

\section{Experiments}

In this section, we systematically carry out extensive experiments to verify the capability of our DDC model. Due to the space restriction we report additional experimental details in the supplement, \textit{e.g.}, network architectures.

\subsection{Datasets}

For a comprehensive comparison, eight popular datasets on image, text and audio, are utilized in experiments. For the image dataset, MNIST \cite{lecun1998gradient}, CIFAR-10 \cite{krizhevsky2009learning}, STL-10 \cite{DBLP:journals/jmlr/CoatesNL11}, ImageNet-10 \cite{Chang2017ICCV}, and ImageNet-Dog \cite{Chang2017ICCV} are used. As for the text datasets, 20NEWS and REUTERS \cite{DBLP:journals/jmlr/LewisYRL04} are used for comparison. An audio dataset AudioSet-20 that randomly chosen from AudioSet \cite{DBLP:conf/icassp/GemmekeEFJLMPR17} is employed. On these datasets, the training and testing patterns of each dataset are jointly utilized in our experiments, as described in \cite{Chang2017ICCV,DBLP:conf/icml/XieGF16,DBLP:conf/cvpr/YangPB16}. Specifically, the term frequency inverse document frequency features\footnote{https://scikit-learn.org/stable} and the Mel-frequency cepstral coefficients\footnote{https://librosa.github.io/librosa/feature.html} are employed to encode texts and audios, respectively. In summary, the number of clusters, and dimensions of patterns on each datasets are listed in Table~\ref{dataset_description}.

\subsection{Evaluation Metrics}

There are three frequently-used metrics for evaluating the clustering results: Normalized Mutual Information (NMI) \cite{DBLP:journals/jmlr/StrehlG02}, Adjusted Rand Index (ARI) \cite{Hubert1985Comparing}, and clustering Accuracy (ACC) \cite{DBLP:conf/icdm/LiD06}. Intrinsically, these metrics range in $[0,1]$, and higher scores always support that more accurate clustering results are achieved.

\subsection{Compared Methods}

Several existing clustering methods are employed to compare with our approach. Specifically, the traditional methods, including Kmeans \cite{DBLP:journals/tkde/WangWSXSL15}, SC \cite{DBLP:conf/nips/Zelnik-ManorP04} and AC \cite{DBLP:journals/pami/FrantiVH06} are adopted for comparison. For the representation-based clustering approaches, as described in \cite{DBLP:conf/icml/XieGF16}, we employ some unsupervised learning methods, including AE \cite{DBLP:conf/nips/BengioLPL06}, SAE \cite{DBLP:conf/nips/BengioLPL06}, DAE \cite{DBLP:journals/jmlr/VincentLLBM10}, DeCNN \cite{DBLP:conf/cvpr/ZeilerKTF10}, and SWWAE \cite{DBLP:journals/corr/ZhaoMGL15}, to learn feature representations and use Kmeans \cite{DBLP:journals/tkde/WangWSXSL15} to cluster data as a post processing. For a comprehensive comparison, recent single-stage methods, including CatGAN \cite{DBLP:journals/corr/Springenberg15}, GMVAE \cite{DBLP:journals/corr/DilokthanakulMG16}, DAC \cite{Chang2017ICCV}, DAC* \cite{Chang2017ICCV}, JULE-SF \cite{DBLP:conf/cvpr/YangPB16}, JULE-RC \cite{DBLP:conf/cvpr/YangPB16}, and DEC \cite{DBLP:conf/icml/XieGF16} are employed.

\begin{table}[t]\small
\renewcommand{\arraystretch}{1}
\centering
\begin{threeparttable}
\caption{ The experimental datasets.}
\label{dataset_description}
\begin{tabular}{L{2.6cm} C{1.3cm} C{1.2cm} C{1.5cm}}
\toprule
Dataset & Numbers & Clusters & Dimensions\\
\midrule
MNIST \cite{lecun1998gradient} & $70000$ & $10$ & $784$\\
CIFAR-10 \cite{krizhevsky2009learning} & $60000$ & $10$ & $3072$\\
STL-10 \cite{DBLP:journals/jmlr/CoatesNL11} & $13000$ & $10$ & $27648$\\
ImageNet-10 \cite{Chang2017ICCV} & $13000$ & $10$ & $27648$\\
ImageNet-Dog \cite{Chang2017ICCV} & $19500$ & $15$ & $27648$\\
20NEWS \cite{DBLP:journals/jmlr/LewisYRL04} & $18846$ & $20$ & $10000$ \\
REUTERS \cite{DBLP:journals/jmlr/LewisYRL04} & $685071$ & $4$ & 2000 \\
AudioSet-20 \cite{DBLP:conf/icassp/GemmekeEFJLMPR17} & $30000$ & $20$ & $1920$ \\
\bottomrule
\end{tabular}
\end{threeparttable}
\vspace{-0.25cm}
\end{table}

\subsection{Experimental Settings}\label{Experimental Settings}

\begin{table*}[!t]\footnotesize
\renewcommand{\arraystretch}{0.896}
\centering
\begin{threeparttable}
\caption{The clustering results of the methods on the experimental datasets. The best results are highlighted in \textbf{bold}.}
\label{results_qr}
\begin{tabular}{L{2cm} C{0.78cm} C{0.78cm} C{0.78cm} C{0.78cm} C{0.78cm} C{0.78cm} C{0.78cm} C{0.78cm} C{0.78cm} C{0.78cm} C{0.78cm} C{0.78cm}}
\toprule
\toprule
\centering{Dataset} & \multicolumn{3}{c}{MNIST \cite{lecun1998gradient}} & \multicolumn{3}{c}{CIFAR-10 \cite{krizhevsky2009learning}} & \multicolumn{3}{c}{STL-10 \cite{DBLP:journals/jmlr/CoatesNL11}} & \multicolumn{3}{c}{ImageNet-10 \cite{Chang2017ICCV}}\\
\cmidrule(lr){2-4}\cmidrule(lr){5-7}\cmidrule(lr){8-10}\cmidrule(lr){11-13}
\centering{Metric} & NMI & ARI & ACC & NMI & ARI & ACC & NMI & ARI & ACC & NMI & ARI & ACC\\
\midrule
Kmeans \cite{DBLP:journals/tkde/WangWSXSL15} & 0.4997 & 0.3652  & 0.5723  & 0.0871  & 0.0487  & 0.2289  & 0.1245  & 0.0608  & 0.1920  & 0.1186  & 0.0571  & 0.2409\\
SC \cite{DBLP:conf/nips/Zelnik-ManorP04} & 0.6626  & 0.5214  & 0.6958  & 0.1028  & 0.0853  & 0.2467  & 0.0978  & 0.0479  & 0.1588  & 0.1511  & 0.0757  & 0.2740\\
AC \cite{DBLP:journals/pami/FrantiVH06} & 0.6094  & 0.4807  & 0.6953  & 0.1046  & 0.0646  & 0.2275  & 0.2386  & 0.1402  & 0.3322  & 0.1383  & 0.0674  & 0.2420\\
AE \cite{DBLP:conf/nips/BengioLPL06} & 0.7257  & 0.6139  & 0.8123  & 0.2393  & 0.1689  & 0.3135  & 0.2496  & 0.1610  & 0.3030  & 0.2099  & 0.1516  & 0.3170\\
SAE \cite{DBLP:conf/nips/BengioLPL06} & 0.7565  & 0.6393  & 0.8271  & 0.2468  & 0.1555  & 0.2973  & 0.2520  & 0.1605  & 0.3203  & 0.2122  & 0.1740  & 0.3254\\
DAE \cite{DBLP:journals/jmlr/VincentLLBM10} & 0.7563  & 0.6467  & 0.8316  & 0.2506  & 0.1627  & 0.2971  & 0.2242  & 0.1519  & 0.3022  & 0.2064  & 0.1376  & 0.3044\\
DeCNN \cite{DBLP:conf/cvpr/ZeilerKTF10} & 0.7577  & 0.6691  & 0.8179  & 0.2395  & 0.1736  & 0.2820  & 0.2267  & 0.1621  & 0.2988  & 0.1856  & 0.1421  & 0.3130\\
SWWAE \cite{DBLP:journals/corr/ZhaoMGL15} & 0.7360  & 0.6518  & 0.8251  & 0.2330  & 0.1638  & 0.2840  & 0.1962  & 0.1358  & 0.2704  & 0.1761  & 0.1603  & 0.3238\\
CatGAN \cite{DBLP:journals/corr/Springenberg15} & 0.7637  & 0.7360  & 0.8279  & 0.2646  & 0.1757  & 0.3152  & 0.2100  & 0.1390  & 0.2984  & 0.2250  & 0.1571  & 0.3459\\
GMVAE \cite{DBLP:journals/corr/DilokthanakulMG16} & 0.7364  & 0.7129  & 0.8317  & 0.2451  & 0.1674  & 0.2908  & 0.2004  & 0.1464  & 0.2815  & 0.1934  & 0.1683  & 0.3344\\
DAC \cite{Chang2017ICCV} & 0.9351 & 0.9486 & 0.9775 & 0.3959 & 0.3059 & 0.5218 & 0.3656 & 0.2565 & 0.4699 & 0.3944 & 0.3019 & 0.5272\\
DAC* \cite{Chang2017ICCV} & 0.9246 & 0.9406 & 0.9660 & 0.3793 & 0.2802 & 0.4982 & 0.3474 & 0.2351 & 0.4337 & 0.3693 & 0.2837 & 0.5026\\
JULE-SF \cite{DBLP:conf/cvpr/YangPB16} & 0.9063  & 0.9139  & 0.9592  & 0.1919  & 0.1357  & 0.2643  & 0.1754  & 0.1622  & 0.2741  & 0.1597  & 0.1205  & 0.2927\\
JULE-RC \cite{DBLP:conf/cvpr/YangPB16} & 0.9130  & 0.9270  & 0.9640  & 0.1923  & 0.1377  & 0.2715  & 0.1815  & 0.1643  & 0.2769  & 0.1752  & 0.1382  & 0.3004\\
DEC \cite{DBLP:conf/icml/XieGF16} & 0.7716  & 0.7414  & 0.8430  & 0.2568  & 0.1607  & 0.3010  & 0.2760  & 0.1861  & 0.3590  & 0.2819  & 0.2031  & 0.3809\\
\midrule
\textbf{DDC} & \textbf{0.9514 } & \textbf{0.9667 } & \textbf{0.9800 } & \textbf{0.4242 } & \textbf{0.3285 } & \textbf{0.5238 } & \textbf{0.3712 } & \textbf{0.2674 } & \textbf{0.4891 } & \textbf{0.4327 } & \textbf{0.3451 } & \textbf{0.5766}\\
\bottomrule
\end{tabular}
\vspace{0.035cm}
\begin{tabular}{L{2cm} C{0.78cm} C{0.78cm} C{0.78cm} C{0.78cm} C{0.78cm} C{0.78cm} C{0.78cm} C{0.78cm} C{0.78cm} C{0.78cm} C{0.78cm} C{0.78cm}}
\toprule
\centering{Dataset} & \multicolumn{3}{c}{ImageNet-Dog \cite{Chang2017ICCV}} & \multicolumn{3}{c}{20NEWS \cite{DBLP:journals/jmlr/LewisYRL04}} & \multicolumn{3}{c}{REUTERS \cite{DBLP:journals/jmlr/LewisYRL04}} & \multicolumn{3}{c}{AudioSet-20 \cite{DBLP:conf/icassp/GemmekeEFJLMPR17}}\\
\cmidrule(lr){2-4}\cmidrule(lr){5-7}\cmidrule(lr){8-10}\cmidrule(lr){11-13}
\centering{Metric} & NMI & ARI & ACC & NMI & ARI & ACC & NMI & ARI & ACC & NMI & ARI & ACC\\
\midrule
Kmeans \cite{DBLP:journals/tkde/WangWSXSL15} & 0.0548  & 0.0204  & 0.1054  & 0.2154  & 0.0826  & 0.2328  & 0.3275  & 0.2593  & 0.5329  & 0.1901  & 0.0780  & 0.2074\\
SC \cite{DBLP:conf/nips/Zelnik-ManorP04} & 0.0383  & 0.0133  & 0.1111  & 0.2147  & 0.0934  & 0.2486  & N/A & N/A & N/A & 0.1774  & 0.0741  & 0.2195\\
AC \cite{DBLP:journals/pami/FrantiVH06} & 0.0368  & 0.0207  & 0.1385  & 0.2024  & 0.0963  & 0.2397  & N/A & N/A & N/A & 0.1854  & 0.0643  & 0.2286\\
AE \cite{DBLP:conf/nips/BengioLPL06} & 0.1039  & 0.0728  & 0.1851  & 0.4439  & 0.3237  & 0.4907  & 0.3564  & 0.3164  & 0.7197  & 0.1933  & 0.1000  & 0.2300\\
SAE \cite{DBLP:conf/nips/BengioLPL06} & 0.1129  & 0.0729  & 0.1830  & 0.4575  & 0.3265  & 0.4869  & 0.3556  & 0.3186  & 0.7256  & 0.1925  & 0.1061  & 0.2382\\
DAE \cite{DBLP:journals/jmlr/VincentLLBM10} & 0.1043  & 0.0779  & 0.1903  & 0.4558  & 0.3274  & 0.5027  & 0.3675  & 0.3386  & 0.7246  & 0.1964  & 0.1048  & 0.2366\\
DeCNN \cite{DBLP:conf/cvpr/ZeilerKTF10} & 0.0983  & 0.0732  & 0.1747  & 0.4450  & 0.3526  & 0.5199  & 0.3750  & 0.3497  & 0.7221  & 0.2028  & 0.1313  & 0.2465\\
SWWAE \cite{DBLP:journals/corr/ZhaoMGL15} & 0.0936  & 0.0760  & 0.1585  & 0.4646  & 0.3247  & 0.5103  & 0.3805  & 0.3538  & 0.7284  & 0.1991  & 0.1323  & 0.2261\\
CatGAN \cite{DBLP:journals/corr/Springenberg15} & 0.1213  & 0.0776  & 0.1738  & 0.4064  & 0.3024  & 0.4754  & 0.3553  & 0.3246  & 0.6245  & 0.2102  & 0.0946  & 0.2174\\
GMVAE \cite{DBLP:journals/corr/DilokthanakulMG16} & 0.1074  & 0.0786  & 0.1788  & 0.4266  & 0.3356  & 0.4792  & 0.3823  & 0.3475  & 0.6365  & 0.2224  & 0.1066  & 0.2441\\
DAC \cite{Chang2017ICCV} & 0.2185 & 0.1105 & 0.2748 & 0.5893 & 0.5005 & 0.5841 & 0.7116 & 0.5892 & 0.7875 & 0.2730 & 0.1372 & 0.2832\\
DAC* \cite{Chang2017ICCV} & 0.1815 & 0.0953 & 0.2455 & 0.5655 & 0.4774 & 0.5495 & 0.6835 & 0.5537 & 0.7535 & 0.2556 & 0.1223 & 0.2645\\
JULE-SF \cite{DBLP:conf/cvpr/YangPB16} & 0.1213  & 0.0776  & 0.1738  & 0.4976  & 0.4506  & 0.5689  & 0.3986  & 0.3594  & 0.6111  & 0.1935  & 0.1128  & 0.2163\\
JULE-RC \cite{DBLP:conf/cvpr/YangPB16} & 0.0492  & 0.0261  & 0.1115  & 0.5126  & 0.4679  & 0.5834  & 0.4012  & 0.3825  & 0.6336  & 0.2106  & 0.1108  & 0.2336\\
DEC \cite{DBLP:conf/icml/XieGF16} & 0.1216  & 0.0788  & 0.1949  & 0.5286  & 0.4683  & 0.5832  & 0.6832  & 0.5635  & 0.7563  & 0.2280  & 0.1102  & 0.2461\\
\midrule
\textbf{DDC} & \textbf{0.2395}&\textbf{0.1283}&\textbf{0.3063}&\textbf{0.6083}&\textbf{0.5247}&\textbf{0.6470}&\textbf{0.7328}&\textbf{0.6044}&\textbf{0.8277}& \textbf{0.2967}&\textbf{0.1744}&\textbf{0.3060}\\
\bottomrule
\bottomrule
\end{tabular}
\end{threeparttable}
\vspace{-0.25cm}
\end{table*}

In our experiments, DNNs with the constraint layer are devised to learn indicator features (the details of the devised DNNs can be found in the supplementary material). Specifically, we update parameters in models on mini batches with $m=1000$ patterns, relying on the traditional spectral clustering \cite{DBLP:conf/nips/Zelnik-ManorP04}. The normalized Gaussian initialization strategy \cite{DBLP:conf/iccv/HeZRS15} is utilized to initialize the devised deep networks. The RMSProp optimizer \cite{Tieleman2012} where the initial learning rate is set to $0.001$ is utilized to train the networks in DDC. For a reasonable evaluation, we perform $10$ random restarts for all experiments and the averages are employed to compare with the others methods.

\subsection{Clustering Results}

The clustering results of the compared methods are listed in Table~\ref{results_qr}. In the table, DDC achieves the best performance on all the datasets (including images, texts and audios), with significant gains. It demonstrates that DDC is effective for managing general clustering tasks in practice. Furthermore, several tendencies can be observed from Table~\ref{results_qr}.

First, the representation learning is assuredly crucial in clustering. From the table, the traditional methods (\textit{e.g.}, Kmeans \cite{DBLP:journals/tkde/WangWSXSL15}) always achieve inferior performance than the representation-based clustering methods (\textit{e.g.}, AE \cite{DBLP:conf/nips/BengioLPL06}). Since the only difference between these methods is the utilized features, the performance gains demonstrate that the feature learning plays an important role in clustering.

Second, combing the representation learning and the traditional methods is beneficial to clustering. From the table, the approaches (\textit{e.g.}, DEC \cite{DBLP:conf/icml/XieGF16}, JULE \cite{DBLP:conf/cvpr/YangPB16}) that can simultaneously perform the feature learning and clustering possess superior performance than the representation-based approaches. This scenario powerfully verifies our motivation in DDC that the representation learning and the clustering are both stimulative to each other.

Finally, DDC suffices to handle large-scale datasets (\textit{e.g.}, ImageNet-10, REUTERS and AudioSet-20), not limited to simple datasets (\textit{e.g.}, MNIST). Specifically, ``large" consists of two aspects, \textit{i.e.}, the dimension of patterns and the number of patterns. For the dimension, DDC enables to cluster high dimensional patterns with DNNs, which can partially manage the curse of dimensionality. As for the number, DDC optimizes DNNs in a mini-batch based optimization, which is independent of the number of patterns and enhances the practicability consequently.

\subsection{Ablation Study}

In this subsection, we carry out extensive ablation studies to synthetically analyze the developed DDC. Intuitively, all the results are orderly illustrated in Figure \ref{ablation}.

\subsubsection{Unobservable Number of Clusters}

\begin{figure*}[!t]
\centering
\centerline{\includegraphics[width=17cm]{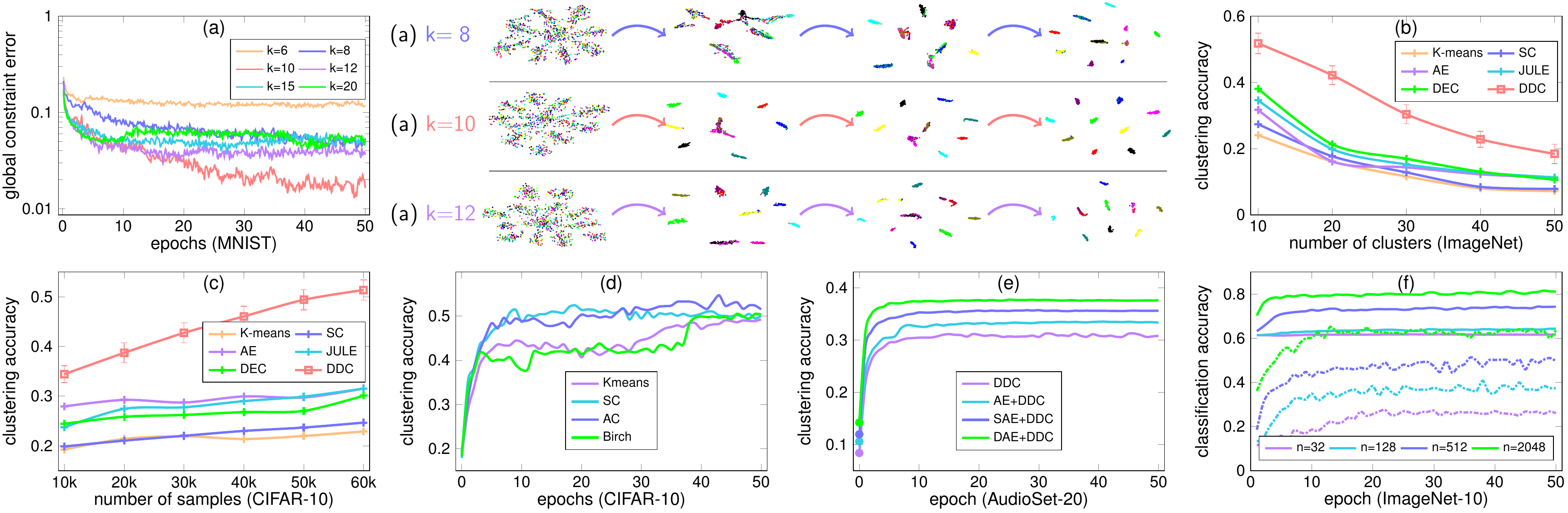}}
\caption{
Ablation studies in DDC. (a) Unobservable number of clusters. Specifically, $k$ indicates the dimensionality of indicator features in DDC, and the clustering results ($k\in\{8,10,12\}$) at different stages are orderly illustrated on the right side. (b) Performance on larger number of clusters. (c) Impact of number of patterns. (d) Validity of learned indicator features. (e) Sensitivity to initializations, where solid circles illustrate the different initial states. (f) Serviceability of learned DNNs, where $n$ means the number of labeled patterns. For each $n$, the solid and dashed lines correspond to the pretrained and un-pretrained DNNs, respectively.
}\medskip
\vspace{-0.45cm}
\label{ablation}
\end{figure*}

An experiment on MNIST is performed to demonstrate the validity of DDC, when the number of clusters $k$ is unobservable. In the basic network on MNIST, we select $k$ from $\{6,8,10,12,15,20\}$ to investigate the variations of global constraint in DDC. From Figure~\ref{ablation} (a), there are obvious differences for different $k$. Specifically, smaller global constraint errors are achieved, as $k$ increases from $6$ to $10$. For the cases when $k>10$, larger global constraint errors are obtained as $k$ gradually increases. When $k=10$, the smallest global constraint error on MNIST is achieved. That is, the global constraint error can be employed to search an appropriate number of clusters. This verifies that DDC enables to handle the clustering tasks when the number of clusters is unobservable. In Figure~\ref{ablation}, the learned indicator features are orderly visualized with t-SNE \cite{maaten2008visualizing} when $k\in\{8,10,12\}$. It indicates that the real number of clusters also corresponds to the fastest convergence rate.

\subsubsection{Performance on Larger Number of Clusters}

We establish five different datasets to evaluate the performance of DDC, when the number of clusters is very large. By varying the number of clusters between $10$ and $50$ with an interval $10$, $5$ subsets of ILSVRC2012-1K \cite{DBLP:conf/cvpr/DengDSLL009} are randomly sampled. Specifically, we observe the following two tendencies from Figure~\ref{ablation} (b). First, the performance of all the methods gradually degrade as the number of clusters increases. Second, the evident superiority always holds on every dataset, which signifies that DDC is in a position to handle the clustering tasks with various number of clusters.

\subsubsection{Impact of Number of Patterns}

Akin to deep supervised learning tasks, we consider to study the impact of number of patterns on our DDC model. To this end, we vary the number of patterns in the CIFAR-10 dataset between $10k$ and $60k$ with an interval $10k$. From the clustering results illustrated in Figure~\ref{ablation} (c), one can observe that the performance of DDC is gradually enhanced as the number of patterns increases. This is reasonable since deep networks always possess a large number of learnable parameters. By utilizing more data to train the networks, these parameters can be estimated more accurately. Consequently, more stronger deep networks are learned. This observation also indicates a possibility that DDC can improve the performance by increasing the amount of data. Because of the accessibility of unlabeled data in practice, there is still much room for improvement.

\subsubsection{Validity of Learned Indicator Features}

To reveal the validity of the learned indicator features in DDC, we report the clustering results which are obtained with the learned indicator features and four popular traditional clustering methods, \textit{i.e.}, Kmeans \cite{DBLP:journals/tkde/WangWSXSL15}, SC \cite{DBLP:conf/nips/Zelnik-ManorP04}, AC \cite{DBLP:journals/pami/FrantiVH06}, and Birch \cite{DBLP:journals/jmlr/StrehlG02}. From Figure~\ref{ablation} (d), almost the same clustering processes are generated for all traditional clustering methods. The slight mismatching may originate from the randomness in training, including random mini-batch selections or initializations. This scenario demonstrates that DDC is capable of learning high-level discriminative representations for clustering and alleviating the influences of these clustering methods consequently.

\subsubsection{Sensitivity to Initializations}

As the previous studies have proved \cite{DBLP:conf/nips/BengioLPL06}, the initial state is crucial for deep networks. Therefore we study the contribution of initializations for DDC. Typically, four frequently-used initial strategies are utilized in this experiment, \textit{i.e.}, the normalized Gaussian initialization strategy \cite{DBLP:conf/iccv/HeZRS15}, AE \cite{DBLP:conf/nips/BengioLPL06}, SAE \cite{DBLP:conf/nips/BengioLPL06} and DAE \cite{DBLP:journals/jmlr/VincentLLBM10}. Intuitively, Figure~\ref{ablation} (e) illustrates the clustering processes on AudioSet-20 with diverse initializations, indicating that the better initializations are beneficial to our DDC model, which is predictable, since preferable initializations can generate proper indicator features and converge to better partitions consequently.

\subsubsection{Serviceability of Learned DNNs}

In this experiment, the serviceability of learned DNNs in DDC is validated. Naturally, clustering with DDC can be treated as a process to train deep networks in an unsupervised manner purely. To prove the validity of the trained networks, we fine-tune the networks with a small number of labeled patterns (\textit{i.e.}, 32, 128, 512, 2048). From Figure~\ref{ablation} (f), there are dramatic margins between the pretrained networks (solid lines) and un-pretrained networks (dashed lines), especially when the labeled patterns are very limited, \textit{e.g.}, 32 (purple lines). These results strongly demonstrate that DDC is capable of training deep networks from random states to informative states in an unsupervised way.

\section{Conclusion}

We develop Deep Discriminative Clustering (DDC) to yield a unified mechanism of learning representations and clustering in a purely data-driven manner. For this purpose, the global and local constraints are introduced, which can be employed to guide the estimation of the relationships between patterns and the learning of discriminative representations. By optimizing the DDC model in a mini-batch way, DDC theoretically converges and the network in DDC can directly generate clustering centers for clustering. Extensive experimental results strongly demonstrate that DDC outperforms current methods on popular image, text and audio datasets concurrently. In the future, a potential direction is to combine DDC and the graph networks \cite{DBLP:journals/corr/abs-1806-01261} to explore high-level relationships for improving the performance.

{\small
\bibliographystyle{ieee}
\bibliography{ddc}
}

\end{document}